\documentclass[sigconf]{acmart}
\AtBeginDocument{%
  \providecommand\BibTeX{{%
    \normalfont B\kern-0.5em{\scshape i\kern-0.25em b}\kern-0.8em\TeX}}}

\usepackage{multirow}
\usepackage{enumitem}
\usepackage{xcolor}
\usepackage{algorithmic}
\usepackage{caption}
\usepackage{amsfonts}
\usepackage{bm}
\usepackage{subcaption}
\usepackage[linesnumbered, ruled]{algorithm2e}

\newcommand{\facc}[2]{{#1}{\scriptsize±{#2}}}

\DeclareMathOperator*{\argmin}{arg\,min}
\newcommand{{\method}}{CAF}
\newcommand{\indep}{\perp \!\!\! \perp}
\definecolor{atomictangerine}{rgb}{1.0, 0.6, 0.4}

\copyrightyear{2023} 
\acmYear{2023} 
\setcopyright{acmlicensed}\acmConference[CIKM '23]{Proceedings of the 32nd ACM International Conference on Information and Knowledge Management}{October 21--25, 2023}{Birmingham, United Kingdom}
\acmBooktitle{Proceedings of the 32nd ACM International Conference on Information and Knowledge Management (CIKM '23), October 21--25, 2023, Birmingham, United Kingdom}
\acmPrice{15.00}
\acmDOI{10.1145/3583780.3615092}
\acmISBN{979-8-4007-0124-5/23/10}

\begin{document}

\title{Towards Fair Graph Neural Networks via Graph Counterfactual}

\author{Zhimeng Guo}
\affiliation{%
  \institution{The Pennsylvania State University}
  \country{USA}
  }
\email{zzg5107@psu.edu}

\author{Jialiang Li}
\affiliation{%
  \institution{New Jersey Institute of Technology}
  \country{USA}
  }
\email{jl2356@njit.edu}

\author{Teng Xiao}
\affiliation{%
  \institution{The Pennsylvania State University}
  \country{USA}
  }
\email{tengxiao@psu.edu}

\author{Yao Ma}
\affiliation{%
  \institution{Rensselaer Polytechnic Institute}
  \country{USA}
  }
\email{may13@rpi.edu}

\author{Suhang Wang}
\authornote{Corresponding Author}
\affiliation{%
  \institution{The Pennsylvania State University}
  \country{USA}
  }
\email{szw494@psu.edu}

\renewcommand{\shortauthors}{Zhimeng Guo, Jialiang Li, Teng Xiao, Yao Ma, and Suhang Wang}

\begin{abstract}

Graph neural networks have shown great ability in representation (GNNs) learning on graphs, facilitating various tasks. Despite their great performance in modeling graphs, recent works show that GNNs tend to inherit and amplify the bias from training data, causing concerns of the adoption of GNNs in high-stake scenarios. Hence, many efforts have been taken for fairness-aware GNNs. However, most existing fair GNNs learn fair node representations by adopting statistical fairness notions, which may fail to alleviate bias in the presence of statistical anomalies. Motivated by causal theory, there are several attempts utilizing graph counterfactual fairness to mitigate root causes of unfairness. However, these methods suffer from non-realistic counterfactuals obtained by perturbation or generation. In this paper, we take a causal view on fair graph learning problem. Guided by the casual analysis, we propose a novel framework {\method}, which can select counterfactuals from training data to avoid non-realistic counterfactuals and adopt selected counterfactuals to learn fair node representations for node classification task. Extensive experiments on synthetic and real-world datasets show the effectiveness of {\method}. Our code is available at~\url{https://github.com/TimeLovercc/CAF-GNN}.

\end{abstract}



\maketitle

\section{Introduction}

Graphs are pervasive in real-world, such as knowledge graphs~\citep{li2022house}, social networks~\citep{kumar2022influence} and biological networks~\citep{liu2022spherical}. Recently, graph neural networks (GNNs)~\citep{kipf2017semisupervised,wu2020a} have shown great ability in modeling graph-structural data. Generally, GNNs adopt the message passing mechanism, which updates a node's representation by iteratively aggregating its neighbors' representations. The resulting representation preserves both node attributes and local graph structure information, facilitating various downstream tasks such as node classification~\cite{kipf2017semisupervised, hamilton2017inductive} and link prediction~\cite{zhang2018link}. 
Despite their great performance, recent studies~\cite{dai2021say,ma2022learning,kose2021fairnessaware} show that GNNs tend to inherit bias from training data, which may result in biased predictions towards sensitive attributes, such as age, gender and race. In addition, the message passing mechanism of GNNs and graph structure could magnify the bias~\cite{dai2021say}. For example, in social networks, nodes of the same race are more likely to connect to each other.  The message passing of GNNs would make the representation of linked nodes similar, resulting in a high correlation of node representation with race, hence the biased prediction.  The biased prediction has raised concerns from ethical and societal perspectives, which severely limits the adoption of GNNs in high-stake decision-making systems~\cite{xiao2021a}, such as job applicants ranking~\citep{mehrabi2021a} and criminal prediction~\citep{jin2020addressing}. 

Hence, many efforts have been taken for fair GNNs~\citep{ma2022learning, kose2021fairnessaware, dai2021say, rahman2019fairwalk}. However, most existing methods are based on statistical fairness notions, which aim to make statistically fair predictions for different sub-groups or individuals~\citep{ loveland2022fairedit}. Several works have pointed out such fairness notions fail to detect discrimination in the presence of statistical anomalies~\citep{makhlouf2020survey, kusner2017counterfactual}. Therefore, there has been a recent shift toward counterfactual fairness in graph modeling~\citep{agarwal2021towards, ma2022learning}. This approach aims to eradicate the root causes of bias by mapping the causal relationships among variables. The identified causal structure allows for the adjustment of sensitive data to generate counterfactuals, ensuring that the prediction remains unaltered by the sensitive information through the utilization of these counterfactuals.
For example, NIFTY~\citep{agarwal2021towards} perturbs sensitive attributes to obtain counterfactuals and maximizes the similarity between original representations and perturbed representations to make representations invariant to sensitive attributes. GEAR~\citep{ma2022learning} adopts GraphVAE~\citep{kipf2016variational} to generate counterfactuals and minimizes the discrepancy between original representations and counterfactual representations to get rid of the influence of sensitive attributes. Despite their superior performance, existing graph counterfactual fairness works need to flip sensitive attributes or generate counterfactuals with GraphVAE, which can easily result in non-realistic counterfactuals.
Such non-realistic counterfactuals may disrupt the underlying latent semantic structure, thereby potentially undermining the model's performance.
This is because simply flipping sensitive attributes cannot model the influence on other features or graph structure causally caused by sensitive attributes~\citep{agarwal2021towards}, and the generative approach lacks supervision of real counterfactuals and could be over-complicated~\citep{ma2022learning}.

Motivated by the discussion above, in this paper, we investigate whether one can obtain counterfactuals within the training data. For example, if a female applicant was rejected by a college, we aim to find another male applicant who has a similar background as the counterfactual applicant. Thus, we can get realistic counterfactuals and avoid the ill-supervised generation process. To achieve our goal, we are faced with several challenges: (i) Graph data is quite complex, thus it is infeasible to directly find counterfactuals in the original data space. Besides, some guidance or rules are needed to find the counterfactuals. (ii) To achieve graph counterfactual fairness, learned representation should be invariant to sensitive attributes and information causally influenced by sensitive attributes. It is critical to design proper supervision to help models get rid of sensitive information. To tackle the aforementioned challenges, we propose a casual view of the graph, label and sensitive attribute. The causal interpretation guides us to find counterfactuals and learn disentangled representations, where the disentangled content representations are informative to the labels and invariant to the sensitive attributes.
Guided by the causal analysis, we propose a novel framework, \textbf{C}ounterfactual \textbf{A}ugmented \textbf{F}air GNN ({\method}), to simultaneously learn fair node representations for graph counterfactual fairness and keep the performance on node classification tasks. 
Specifically, based on the causal interpretation, we derive several constraints to enforce the learned representations being invariant across different sensitive attributes. To obtain proper counterfactuals to guide representation learning, we utilize labels and sensitive attributes as guidance to filter out potential counterfactuals in representation space. Our main contributions are:
\begin{itemize}[leftmargin=*]
    \item 
    We provide a causal formulation of the fair graph learning process and fair node representation learning task.
    \item We propose a novel framework {\method} to learn node representations for graph counterfactual fairness. Specifically, we find counterfactuals in representation space and design novel constraints to learn the content representations. 
    \item %
    We conduct extensive experiments on real-world datasets and synthetic dataset to show the effectiveness of our model.
\end{itemize}

\section{Related Works}

\noindent\textbf{Graph Neural Networks.} Graph neural networks (GNNs) have dominated various tasks on graph-structured data, such as node classification~\citep{kipf2017semisupervised, velickovic2018graph, xiao2021learning, xiao2022decoupled, chen2022bagnn}, graph classification~\citep{sui2022causal} and link prediction~\citep{zhang2018link,zhao2022learning}. Existing GNNs can be categorized into spatial-based GNNs and spectral-based GNNs. Spatial-based GNNs leverage the graph structure directly, focusing on the relationships between nodes and their immediate neighbors to inform feature learning. On the other hand, spectral-based GNNs operate in the spectral domain defined by the graph Laplacian and its eigenvectors, making them better suited to capture global properties of the graph. 
The superior performance of GNNs has greatly extended their application scenarios~\citep{hamilton2020graph}. For example, banks may leverage GNNs to process transaction networks to detect the abnormal behavior of users~\citep{dai2022a}. The applications in critical decision-making systems place higher requirements for GNNs, such as being fair and interpretable~\citep{yuan2022explainability}. 
Despite their extensive utility and efficacy, recent studies~\cite{kose2021fairnessaware, dai2021say, dai2023unified} show that GNNs can harbor implicit biases on different groups, which can lead to skewed or unfair outcomes. This bias issue is particularly critical when GNNs are deployed in high-stake scenarios, making it necessary to ensure fairness in the modeling process~\cite{madras2018learning}. Thus, mitigating bias and promoting fairness in GNNs are active and necessary research areas~\citep{dai2022a}.
The source of bias in Graph Neural Networks (GNNs) primarily originates from two areas. First, it comes from the inherent bias in the input data, which may contain unequal representation or prejudiced information about nodes or connections in the graph. Second, the bias can stem from the algorithmic design of the GNN itself, which may unintentionally emphasize certain features or connections over others during the learning process. Therefore, there is a trend for the research community to design fairer GNN models to deal with graph-based tasks~\citep{jiang2022fmp, loveland2022fairedit,dai2022a, dai2021say}. 

\noindent\textbf{Fairness in GNNs.} Fairness is a widely-existed issue of machine learning systems~\citep{sekhon2016perceptions, madras2018learning, mehrabi2021a, zhu2023fairness}. Researchers evaluate the fairness of models with many kinds of fairness notions, including group fairness~\citep{hardt2016equality, zemel2013learning}, individual fairness~\citep{dwork2012fairness} and counterfactual fairness~\citep{kusner2017counterfactual}. The metrics can also be used to measure the fairness performance of Graph Neural Networks~\citep{agarwal2021towards, ma2022learning}. The commonly used fairness notions in GNNs are statistical parity~\citep{zemel2013learning} and equal opportunity~\citep{hardt2016equality}. FairGNN~\citep{dai2021say} utilizes adversarial training to establish fairness in graph-based models, refining its representation through an adversary tasked with predicting sensitive attributes. EDITS~\citep{dong2022edits}, on the other hand, is a pre-processing technique that focuses on ensuring fairness in graph learning. It aims to eliminate sensitive information from the graph data by correcting any inherent biases present within the input network.
However, these methods and their metrics are developed based on correlation~\citep{makhlouf2020survey}, which has been found to be unable to deal with statistical anomalies, such as Simpson's paradox~\citep{simpson1951the}. Based on the causal theory, counterfactual fairness can model the causal relationships and gets rid of the correlation-induced abnormal behavior~\citep{makhlouf2020survey, kusner2017counterfactual}. There is an increasing interest to apply counterfactual fairness on graphs to design fairer GNNs~\citep{agarwal2021towards, ma2022learning}. NIFTY~\citep{agarwal2021towards} perturbs sensitive attributes for each node to obtain counterfactuals and omits the causal relationships among variables. GEAR~\citep{ma2022learning} uses GraphVAE~\citep{kipf2016variational} to generate the graph structure and node features causally caused by the sensitive attributes.
For more details about counterfactual learning on graphs, please refer to the survey~\cite{guo2023counterfactual}.

Our paper is inherently different from existing work: (i) Unlike existing works that might generate unrealistic counterfactuals, our work avoids the generation process and selects counterfactuals with sensitive attributes and labels as guidance; and (ii) We propose a causal view to understand the source of bias. Based on the causal interpretation, we also design several constraints to help our model learn the fair node representations.

\section{Preliminaries}

In this section, we start by introducing the necessary notation and defining the problem at hand. Following this, we employ the Structural Causal Model to frame the issue, which will then motivate our solution - the disentangled fair representation learning method.

\label{preliminaries}

\subsection{Notations and Problem Definition}
Throughout the paper,  we use italicized uppercase letters to represent random variables (e.g., $S$, $E$) and use italicized lowercase letters to denote the specific value of scalars (e.g., $s$, $y_i$). Non-italicized bold lowercase and uppercase letters are used to denote specific values of vectors (e.g., $\mathbf{x}_i$) and matrices (e.g., $\mathbf{X}$), respectively.

Let $\mathcal{G}=(\mathcal{V}, \mathcal{E}, \mathbf{X})$ denote an attributed graph, where $\mathcal{V}=\{v_{1}, ..., v_{N}\}$ is the set of $N$ nodes, $\mathcal{E} \subseteq \mathcal{V} \times \mathcal{V}$ is the set of edges, $\mathbf{X} \in \mathbb{R}^{N \times D}$ is the node attribute matrix. The $i$-th row of $\mathbf{X}$, i.e., $\mathbf{x}_i$ is the feature vector of node $v_i$. $\mathbf{A} \in \{0,1\}^{N \times N}$ is the adjacency matrix of the graph $\mathcal{G}$, where $\mathbf{A}_{ij}=1$ if nodes $v_i$ and $v_j$ are connected; otherwise $\mathbf{A}_{ij}=0$. We use $\mathbf{s} \in \{0,1\}^{N \times 1}$ to denote the sensitive attributes, where $s_i$ is the sensitive attribute of $v_i$. 
Following~\citep{ma2022learning}, we only consider binary sensitive attributes and leave the extension of multi-category sensitive attributes as future work. We use $\mathbf{y} \in \{1,...,c\}^{N\times 1}$ to denote the ground-truth node labels, where $y_i$ is the label of $v_i$. In this paper, we assume that both target labels and sensitive attributes are binary variables for convenience. 

For the semi-supervised node classification task, only part of nodes $\mathcal{V}_L \in \mathcal{V}$ are labeled for training and the remaining nodes $\mathcal{V}_U=\mathcal{V}\backslash \mathcal{V}_L$ are unlabeled.  The goal is to train a classifier $f$ to predict the labels of unlabeled nodes, which has satisfied node classification performance and fairness performance simultaneously.
Given  $\mathbf{X}$, $\mathbf{A}$ and $\mathbf{Y}_L$, the goal of semi-supervised node classification is to learn a mapping function $f$ to predict the labels of unlabeled nodes, i.e.,  $f: (\mathbf{A}, \mathbf{X}) \rightarrow \mathcal{Y}_U$, where $\mathcal{Y}_U$ the set of predicted labels of unlabeled nodes $\mathcal{V}_U$.

\subsection{The Desiderata for Fair Graph Learning} 
\begin{figure}[t]
    \centering
    \includegraphics[width=0.6\linewidth]{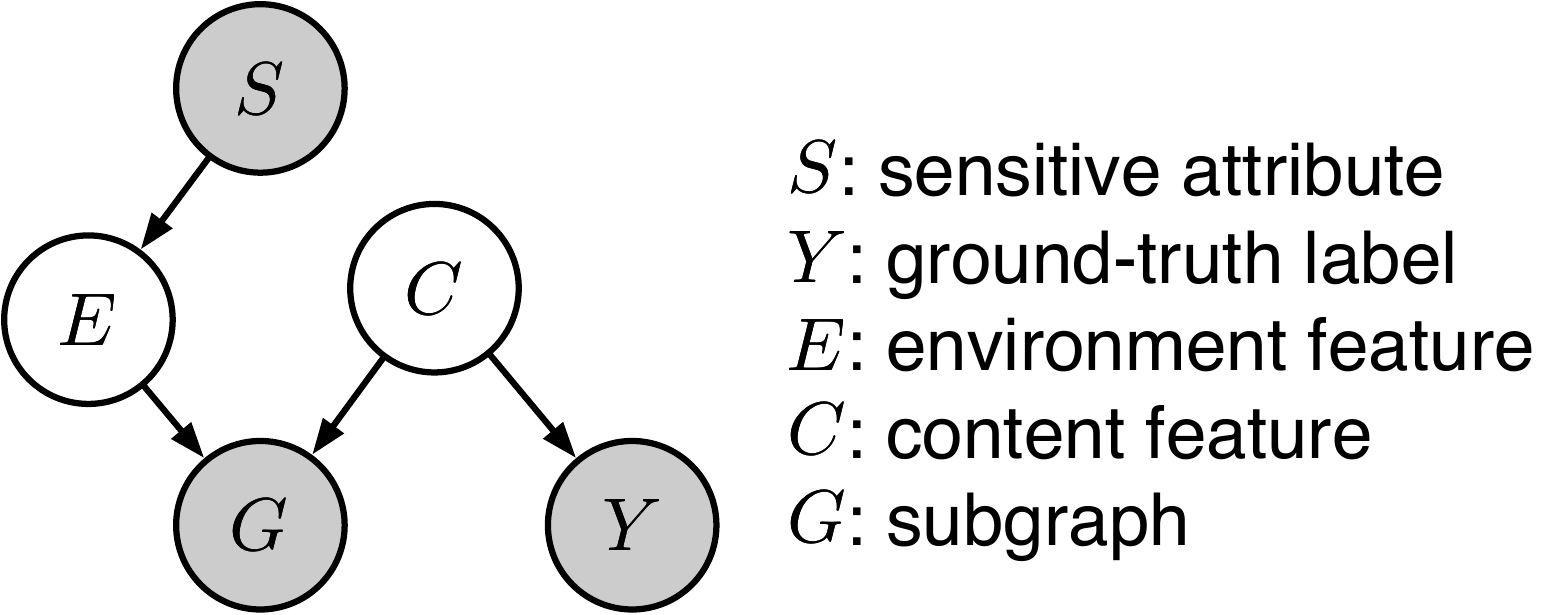}
    \vskip -1em
    \caption{Structural Causal Model for model prediction. We use white color to denote latent variables and use gray color to denote the observed variables.
    }
    \label{fig:scm}
    \vskip -1em
\end{figure}

GNNs have shown remarkable capabilities in the realm of semi-supervised node classification. However, they are not immune to bias issues, primarily stemming from imbalanced or prejudiced input data, and potentially from the structural design of the GNNs themselves, which may inadvertently prioritize certain features or connections. Therefore, substantial efforts have been directed towards developing fairness-aware methodologies within GNNs. The majority of these methods strive to ensure correlation-based fairness notions, such as 
demographic parity or equality of opportunity. However, these correlation-based fairness notions can be inherently flawed, particularly in the presence of statistical anomalies, which calls for more nuanced and robust approaches to achieve fairness in GNNs. Recent advance~\citep{makhlouf2020survey} shows that causal-based fairness notions can help resolve this issue. Thus, to help design a fair GNN classifier, we take a deep causal look under the observed graph. Without loss of generality, in this work, we focus on the node classification task and construct a Structural Causal Model~\citep{shanmugam2018elements} in Figure~\ref{fig:scm}. It presents the causal relationships among five variables: sensitive attribute $S$, ground-truth label $Y$, environment feature $E$, content feature $C$ and ego-graph $G$ for each node. Each link denotes a deterministic causal relationship between two variables. We list the following explanations for the SCM:
\begin{itemize}[leftmargin=*]
    \item $ S \rightarrow E.$ The variable $E$ denotes latent environment features that are determined by the sensitive attribute $S$. For example, people of different genders will have different heights or other physical characteristics, where $S$ is the sensitive attribute of genders and $E$ is physical characteristics that are causally determined by the sensitive attribute. This relationship will lead to bias in latent feature space, which we will explain shortly. 
    \item $ C \rightarrow Y.$ %
    The variable $C$ denotes the content feature that determines ground-truth label $Y$. Taking the credit scoring as an example, ideally, we assign credit scores using personal information not related to the sensitive attribute, i.e., we use content feature $C$ instead of $E$ to assign credit score $Y$.
    \item $E \rightarrow G \leftarrow C.$ The ego-graph $G$ is determined by the content feature $C$ and the environment feature $E$, which are two disjoint parts. $E$ and $C$ are latent features and $G$ is the observed ego-graph. Considering a one-hop ego-graph, it contains the social connections of center node and the observed feature of center node. The causal relationship indicates environment feature $E$ and content feature $C$ can determine one's social connections and personal features (node attributes).
\end{itemize}

The SCM paves us a way to understand the source of bias and how to design a fair GNN classifier. Next, we give details about source of bias and disentangled learning. Our objective is to approximate the content feature $C$ with a content representation denoted as $\hat{C}$, and similarly, approximate the environment feature $E$ with an environment representation denoted as $\hat{E}$. To streamline our discussion, we will slightly abuse notation by also employing the symbols $C$ and $E$ to signify the corresponding content and environment representations throughout the remainder of the paper. 

\subsubsection{Source of Bias}
From the causal graph, we can observe that the sensitive variable $S$ and the label variable $Y$ are independent with each other, i.e., the only path from $S$ to $Y$, $S\rightarrow E \rightarrow G \leftarrow C \leftarrow Y$ is blocked by the collider $G$. However, it is worthy noting that $S$ and $Y$ are \textbf{dependent conditioned on} $G$, i.e., %
\begin{equation}
    P(Y,S|G) \ne P(Y|G) P(S|G).
\end{equation}
The conditional dependency of $Y$ and $S$ on $G$ is one major reason that leads to biased prediction. If we directly learn a GNN model that aims to predict $Y$ based on $G$, as $Y$ and $S$ are dependent given $G$, the learned label $Y$ will have correlation with $S$, resulting in the biased prediction on sensitive attribute $S$.

Alternatively, we can understand the bias by treating existing GNNs as composed of a feature extractor $g$ and a classifier $c$. The feature extractor $g$ takes the subgraph centered at a node as input and learns node representation as $\mathbf{z} = g(G)$. Then the classifier $c$ uses the representation $\mathbf{z}$ to predict the label as $\hat{y} = c(\mathbf{z})$. As $G$ is dependent on $E$ and $C$, the learned representation $\mathbf{z}$ is likely to contain mixed information of both $E$ and $C$. Hence, the predicted label $\hat{y}$ is also likely to have correlation with $S$.

\subsubsection{Disentangled Fair Representation Learning} 
From the above analysis motivates, we can observe that in order to have fair prediction, we need to learn disentangled representation $E$ and $C$ to block the path from $S$ to $Y$ conditioned on $G$, and only use the content information $C$ to predict $Y$, i.e., $P(Y|C)$. As $C$ determines $Y$, it contains all the label information to predict $Y$. 
Meanwhile, observing $E$ and $C$ can block the conditional path from $S$ to $Y$, i.e., $P(Y,S|E,C,G)=P(Y|C,E,G)P(S|C,E,G)$. Note that  observing $C$ blocks the path from $E$ to $Y$ and the path from $G$ to $Y$. Hence, we have $P(Y|C,E,G) = P(Y|C)$. Observing $E$ blocks the path from $S$ to $G$ and the path from $S$ to $C$, thus, we have $P(S|C,E,G)=P(S|E)$. This gives us
\begin{equation}
    P(Y,S|E,C,G)=P(Y|C)P(S|E).
\end{equation}
The above equation shows that observing $E$ and $C$ would make $Y$ and $S$ independent and $P(Y|C)$ is unbiased. 

Hence, if we can learn disentangled latent representation $E$ and $C$, we would be able to use $C$ for fair classification. However, the main challenge is we do not have ground-truth $E$ and $C$ to help us train a model that can learn disentangled representation. 
With a slight abuse of notation, we also use $C$ to denote the learned content representation and use $E$ to denote the learned environment representation. Fortunately, we can use the SCM to derive several properties of the optimal representation, which would be used to help learn the latent representation of $C$ and $E$: 
\begin{itemize}[leftmargin=*]
    \item \textit{Invariance}: $C \indep E.$ This property can be understood in two perspectives. That is, the content representations should be independent to the sensitive attributes and the environment representation induced by the sensitive attribute. Meanwhile, the environment representations should be independent to the labels and the content representation which is informative to the labels.
    \item \textit{Sufficiency}: $(C, E) \rightarrow G.$ The combined representation can used to reconstruct the observed graph.
    \item \textit{Informativeness}: $C \rightarrow Y.$ The content representations should have the capacity to give accurate predictions of labels $Y$. 
\end{itemize}

\section{Methodology}

\begin{figure*}[t]
\centering
    \includegraphics[width=0.96\linewidth]{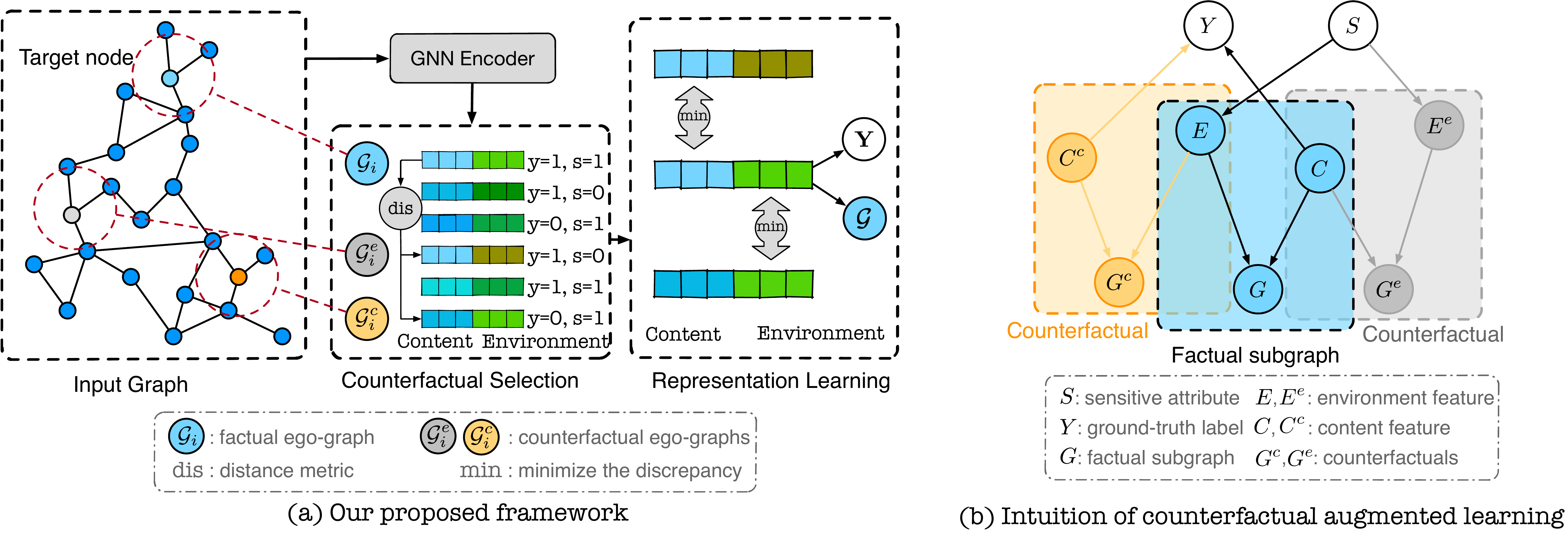}
    \vskip -1.3em
    \caption{An illustration of (a) our proposed framework; (b) intuition of counterfactual augmented learning.  %
    }
    \label{fig:frame}
\end{figure*}

The causal view suggests us to learn disentangled representation $\mathbf{c}$ and $\mathbf{e}$ for node $v$, with $\mathbf{c}$ capturing the content information that is useful for label prediction and irrelevant to sensitive attributes, and $\mathbf{e}$ capturing the environment information depends on sensitive attribute only. With the disentanglement, $\mathbf{c}$ can be used to give fair predictions. However, how to effectively disentangle $\mathbf{c}$ and $\mathbf{e}$ remains a question given that we do not have ground-truth of disentangled representation. Intuitively, for a node $v$ with sensitive attribute $s$, its content representation $\mathbf{c}$ should remain the same when the sensitive attribute is flipped to $1-s$ while its environment representation $\mathbf{e}$ should change correspondingly. Hence, if we know the counterfactual of node $v$, we will be able to utilize the counterfactual to help learn disentangled representation for fair classification; while the counterfactual is not observed. To address the challenges, we propose a novel framework {\method} as shown in Figure~\ref{fig:frame}~(a), which is composed of: (i) a GNN encoder that takes ego-graph $\mathcal{G}$ of node $v$ to learn disentangled representation $\mathbf{c}$ and $\mathbf{e}$; (ii) the counterfactual augmentation module, which aims to discover counterfactual for each factual observation and utilize the counterfactual to help learn disentangled representation; (iii) a fair classifier which takes $\mathbf{c}$ as input for fair classification. Next, we give the details of each component. %

\subsection{Disentangled Representation Learning}
For each node $v_i$, the content representation $\mathbf{c}_i$ should capture the important node attribute and neighborhood information for predicting the label while the environment representation $\mathbf{e}_i$ should capture all important information relevant to sensitive attribute. As GNNs have shown great ability in modeling graph structured data, we adopt GNNs to learn $\mathbf{c}_i$ and $\mathbf{e}_i$. Instead of adopting two GNNs to learn $\mathbf{c}_i$ and $\mathbf{e}_i$ separately, to reduce the number of parameters, we adopt one GNN to learn $\mathbf{c}_i$ and $\mathbf{e}_i$. We empirically found that using two GNNs and one GNN have similar performance due to constraints we designed to disentangle $\mathbf{c}_i$ and $\mathbf{e}_i$, which will be introduced later. Specifically, the GNN $f_{\theta}$ parameterized by $\theta$ takes $\mathcal{G}$ as input and learns representation as:  %
\begin{equation}
\label{repre}
    [\mathbf{C}, \mathbf{E}] = \mathbf{H} = f_{\theta}(\mathbf{A}, \mathbf{X}),
\end{equation}
where $\mathbf{H} \in \mathbb{R}^{N \times d}$ is the learned representation matrix with the $i$-th row, i.e., $\mathbf{h}_i$, as the representation of node $v_i$. We treat the first $d_c$ columns as the content representation matrix $\mathbf{C}$ and use the next $d_e$ columns as the environment representation matrix $\mathbf{E}$. Note that $d = d_c+d_e$. In our implementation, we set $d_c = d_e$. $\mathbf{C} \in \mathbb{R}^{N \times d_c}$ is the content representation matrix with the $i$-th row, i.e., $\mathbf{c}_i$, as the content representation of node $v_i$. Similarly,  $\mathbf{E} \in \mathbb{R}^{N \times d_e}$ is the environment representation matrix with the $i$-the row, i.e., $\mathbf{e}_i$ as the environment representation of node $v_i$. $f_\theta$ is flexible to be various GNNs such as GCN~\citep{kipf2017semisupervised} and GraphSAGE~\citep{hamilton2017inductive}.

 To make sure  $\mathbf{c}_i$ captures the content information for fair label prediction, and $\mathbf{e}_i$ and $\mathbf{c}_i$ are disentangled, based on the causal analysis in Section~\ref{preliminaries}, we add  following constraints:

\textbf{Informativeness Constraint.}
First, the content representation $\mathbf{c}_i$ should be informative to the downstream tasks, i.e., $C \rightarrow Y$. Hence, for node $v_i$, we should be able to get accurate label prediction from $\mathbf{c}_i$. Thus, we introduce a classifier $f_{\phi}$ with model parameter $\phi$. It takes $\mathbf{c}_i$ as input and predicts the class distribution of $v_i$ as:%
\begin{equation}
\label{equ: pred_head}
    \hat{\mathbf{y}}_i=f_\phi(\mathbf{c}_i).
\end{equation}
The loss function for training the classifier is given as:
\begin{equation}
\label{equ: pred}
    \mathcal{L}_{\text{pred}} = \frac{1}{|\mathcal{V}_L|} \sum_{v_i \in \mathcal{V}_L} \ell(\hat{\mathbf{y}}_i, \mathbf{y}_i),
\end{equation}
where $\mathbf{y}_i$ is the one-hot encoding of ground-truth label of $v_i$. $\ell(\hat{\mathbf{y}}_i, \mathbf{y}_i)$ denotes the cross entropy between $\hat{\mathbf{y}}_i$ and $\mathbf{y}_i$.

\textbf{Sufficiency Constraint.} 
As shown in our causal view, the representation $(\mathbf{c}_i$ and $\mathbf{e}_i)$ should be sufficient to reconstruct the observed factual graph $\mathcal{G}_i$. In disentangled representation learning research, the reconstruction supervision is usually adopted to guide the learning process~\citep{yang2021causalvae, higgins2017betavae}. 
However, existing graph counterfactual fairness approaches~\citep{agarwal2021towards, ma2022learning} fail to provide supervision to preserve graph information in the representations. Thus, they put their models at a risk of being stuck in trivial solutions to merely get spurious information in the representations, which contradicts the SCM and is not sufficient to reconstruct the observed graph $\mathcal{G}_i$
. In our model, we formalize the sufficiency constraint as a reconstruction of the graph structure. Specifically, for a pair of nodes $(v_i,v_j)$, we predict the link existence probability as $p_{ij} = \sigma(\mathbf{h}_i \mathbf{h}_j^T)$, where $\mathbf{h}_i = [\mathbf{c}_i, \mathbf{e}_i]$ is the node representation of node $v_i$. The sufficiency constraint is
\begin{equation}
\mathcal{L}_{\text{suf}} = \frac{1}{|\mathcal{E}|+|\mathcal{E}^-|} \sum_{(v_i,v_j) \in \mathcal{E} \cup \mathcal{E}^-} -e_{ij} \log p_{ij} - (1 - e_{ij}) \log p_{ij},
\end{equation}
where $\mathcal{E}^-$ is the set of sampled negative edges. $e_{ij}=1$ if node $v_i$ and $v_j$ are connected; otherwise $e_{ij}=0$.  %

\textbf{Orthogonal Constraint.} The above model can help to learn $\mathbf{c}_i$ that captures graph information for label prediction, however, it doesn't guarantee that $\mathbf{c}_i$ doesn't contain sensitive attribute information. To make sure that $\mathbf{c}_i$ and $\mathbf{e}_i$ are disentangled, i.e., $\mathbf{c}_i$ doesn't contain any environment information relevant to sensitive attribute, we further impose the orthogonal constraint, i.e., $\mathbf{c}_i^T \mathbf{e}_i = 0$.

\subsection{ Counterfactual Augmented Learning}

As we do not have ground-truth of $\mathbf{c}_i$ and $\mathbf{e}_i$, we used several constraints to implicitly supervise the learning of $\mathbf{c}_i$ and $\mathbf{e}_i$. To fully learn disentangled $\mathbf{c}_i$ and $\mathbf{e}_i$, we propose to learn better $\mathbf{e}_i$ and $\mathbf{c}_i$ that follows the counterfactual constraints. As shown in Figure~\ref{fig:frame}~(b), generally, for a node $v_i$ with observe the factual sensitive attribute $s_i$ and label $y_i$, its content representation $\mathbf{c}_i$ should remain similar when the sensitive attribute is flipped to $1-s_i$ but its environment representation $\mathbf{e}_i$ should change correspondingly, which forms the counterfactual subgraph $\mathcal{G}_i^e$.
Similarly, when flip label $y_i$ but keep the sensitive attribute $s_i$ unchanged, then $v_i$'s environment representation $\mathbf{e}_i$ remain the same, while its content representation should change accordingly, leading to the counterfactual subgraph $\mathcal{G}_i^c$. Thus, if we know $\mathcal{G}_i^e$ and $\mathcal{G}_i^c$, we would be able to use these counterfactual graphs together with factual graph $\mathcal{G}_i$ to guide the learning of $\mathbf{c}_i$ and $\mathbf{e}_i$. However, in real-world, we can only observe factual graphs. To solve this challenge, we propose to find potential candidate counterfactuals with the observed factual graphs.

The sensitive attribute and label are used to find counterfactuals in our model. Considering the fair credit scoring problem, when someone was assigned a low score, straightforward thinking is to know the results of people who have a similar background to her but of a different gender.  For example, Sarah, a female, got a low credit score. Then she may ask, what if I were a male, what will my credit score be?
This thinking inspires us to directly find counterfactuals from the observed node samples instead of performing perturbation or generating~\citep{agarwal2021towards, ma2022learning}. The advantages of selecting counterfactuals from the observed node samples are twofold: 
(1) It avoids making assumptions about the graph generation process with sensitive attributes. (2) It does not need additional supervision signal.

Another problem comes: selecting counterfactuals from the original data space is also challenging due to the complexity of graph distance calculation. To get counterfactual $\mathcal{G}^e_i$, we need to find some nodes which have different sensitive attribute and the same label. Similarly, we find some nodes with the same sensitive attribute and different labels as counterfactual $\mathcal{G}^c_i$. The task can be formalized as:
\begin{align}
\mathcal{G}^c_i = \argmin_{\mathcal{G}_j \in \mathbb{G}}\{m(\mathcal{G}_i, \mathcal{G}_j)\ | y_i \neq y_j, s_i = s_j \},
\label{equ:cfs1} \\
\mathcal{G}^e_i = \argmin_{\mathcal{G}_j \in \mathbb{G}}\{m(\mathcal{G}_i, \mathcal{G}_j)\ | y_i = y_j, s_i \neq s_j \},
\label{equ:cfs2}
\end{align}
where $\mathbb{G}=\{\mathcal{G}_i | v_i \in \mathcal{V})\}$ and $m(\cdot, \cdot)$ is a metric of measuring the distance between a pair of subgraphs. Nevertheless, the problem of computing the distance of pairs of graphs is inefficient and infeasible due to the complex graph structure and large search space~\citep{zhao2022learning}. 
As we already have node representations $\mathbf{h}_i = [\mathbf{c}_i,\mathbf{e}_i]$ that capture the graph structure and node attribute information, we propose to measure the distance in the latent space, which can greatly reduce the computation burden.
Then the counterfactual graph searching problem in Eq.~\eqref{equ:cfs1} and Eq.~\eqref{equ:cfs2} is converted to the problem below: 
\begin{align}
    \mathbf{h}^c_i & = \argmin_{\mathbf{h}_j \in \mathbb{H}}\{\|\mathbf{h}_i - \mathbf{h}_j\|_2^2 \ | y_i \neq y_j, s_i = s_j \} \label{equ: select cf1}, \\
     \mathbf{h}^e_i & = \argmin_{\mathbf{h}_j \in \mathbb{H}}\{\|\mathbf{h}_i - \mathbf{h}_j\|_2^2 \ | y_i = y_j, s_i \neq s_j \}, \label{equ: select cf2}
\end{align}
where $\mathbb{H} = \{\mathbf{h}_i | v_i \in \mathcal{V}\}$ and  we use L2 distance to find counterfactuals. A problem is that we only have limited labels in the training set. So we first pre-train the backbone model. With pre-trained model, we can obtain the prediction for unlabeled nodes as pseudo-labels. The pseudo-labels work as the guidance of the counterfactual searching problem.
Note that for each factual input we can also get multiple counterfactuals by selecting a set of counterfactuals in Eq.~\eqref{equ: select cf1} and Eq.~\eqref{equ: select cf2} instead of one. Thus, the counterfactual $\mathcal{G}^c_i$ can be naturally extended to a set of $K$ counterfactuals $\{ \mathcal{G}^{c_k}_i|k=1,...,K\}$ and $\mathcal{G}^e_i$ can be extended to $\{ \mathcal{G}^{e_k}_i|k=1,...,K\}$. We fix $K$ to 10 in our implementation.

We can utilize the counterfactuals to supervise the disentanglement of $\mathbf{c}_i$ and $\mathbf{e}_i$. Specifically, 
as shown in Figure~\ref{fig:frame}~(b), counterfactual $\mathcal{G}^{e_k}_i$ shares the same content information with factual graph $\mathcal{G}_i$ and has different environment information. Without supervision, the factual content representation $\mathbf{c}_i$ and the counterfactual content representation $\mathbf{c}^{e_k}_i$ may contain both the content information and environment information. When we minimize the discrepancy of the learned representations with $\operatorname{dis}(\mathbf{c}_i,\mathbf{c}^{e_k}_i)$, $f_\theta$ will tend to merely keep the content information and squeeze the sensitive information out of learned representations.
In a similar manner, we can use $\operatorname{dis}(\mathbf{e}_i, \mathbf{e}^{c_k}_i)$ to make the environment representation $\mathbf{e}_i$ be invariant to the content information stored in $\mathbf{c}_i$. Also, we put the orthogonal constraint here to encourage $\mathbf{c}_i$ and $\mathbf{e}_i$ to store different information in representation space. The invariance constraint is given as:
\begin{equation}
\begin{aligned}
    \mathcal{L}_{\text{inv}}=\frac{1}{|\mathcal{V}| \cdot K} \sum_{v_i \in \mathcal{V}}\sum_{k = 1}^K  \big[ & \operatorname{dis}(\mathbf{c}_i, \mathbf{c}^{e_k}_i) + \operatorname{dis}(\mathbf{e}_i, \mathbf{e}^{c_k}_i) \\ & + \gamma K\cdot |\operatorname{cos}(\mathbf{c}_i, \mathbf{e}_i)| \big],
\end{aligned}
\end{equation}
where $\operatorname{dis(\cdot, \cdot)}$ is a distance metric, such as the cosine distance and L2 distance in our implementation. $|\operatorname{cos(\cdot, \cdot)}|$ is the absolute value of cosine similarity and we optimize this term to approximate $\mathbf{c}_i^T \mathbf{e}_i=0$. $\gamma$ is the hyper-parameter to control the orthogonal constraint.

\subsection{Final Objective Function of {\method}}
Putting the disentangled representation learning module and the counterfactual selection module together, the final objective function of the proposed {\method} framework is:
\begin{equation}
\label{equ: all}
	\min_{\theta,\phi}\mathcal{L}= \mathcal{L}_{\text {pred }} + \alpha \mathcal{L}_{\text {inv }} + \beta \mathcal{L}_{\text {suf }},
\end{equation}
where $\theta$ and $\phi$ are parameters for the GNN encoder and the prediction head, respectively. $\alpha$ and $\beta$ are hyper-parameters controlling the invariance constraint and the sufficiency constraint.

\subsection{Training Algorithm} 
The whole process of {\method} is summarized in Algorithm~\ref{alg:Framwork}. Our method relies on the counterfactuals in the representation space to guide the disentanglement. However, the randomly initialized representation at the first several epochs may degrade the performance of our model. Therefore, we first pre-train a plain node representation learning model $\mathbf{Y}=g_{\Theta,\Phi}(\mathbf{A}, \mathbf{X})$ only with $\mathcal{L}_{\text {pred }}$. Then we use the optimized parameters $\Theta^*, \Phi^*=\min_{\Theta, \Phi}\mathcal{L}_{\text {pred }}$ to initialize the parameters $\theta$ and $\phi$ of our model and use the aforementioned framework to get the desired disentangled representations. We do not necessarily update the counterfactuals for each epoch. We update the counterfactuals once for $t$ epochs and $t=10$ in our implementation. As shown in Algorithm~\ref{alg:Framwork}, we first pre-train $g_{\Theta, \Phi}$ and use the optimized parameter to initialize $f_\theta$ and $\phi$ from line 1 to line 2. Then we iteratively optimize $f_\theta$ and $\phi$ from line 3 to line 10. In each iteration, we first perform forward propagation to get node representations in line 4. And then for each $t$ epoch we update the selected counterfactuals once from line 5 to line 7. Afterwards, we compute the overall objective and perform backpropagation to optimize the parameters $\theta$ and $\phi$ from line 8 to line 9. After training, we obtain the desired fair model $f_\theta$  and $f_\phi$ in line 11.

\renewcommand{\algorithmicrequire}{\textbf{Input:}}
\renewcommand{\algorithmicensure}{\textbf{Output:}}
\begin{algorithm}[t] 
\caption{ Training Algorithm of {\method}.} 
\label{alg:Framwork} 
\begin{algorithmic}[1]
\REQUIRE
$g=(\mathcal{V},\mathcal{E}, \mathbf{X})$, $\mathbf{Y}_L$, $t$, $\alpha$, $\beta$, $\gamma$, $K$ and $num\_epoch$
\ENSURE $f_\theta$ and $f_\phi$
\STATE Pre-train $g_{\Theta, \Phi}$ based on $\mathcal{L}_{\text{pred}}$ with Eq.~\eqref{equ: pred}
\STATE Use the optimized $\Theta^*$ and $\Phi^*$ to initialize $f_\theta$ and $f_\phi$
\FOR{\text{epoch in range($num\_epoch$)}} 
\STATE Compute representations $\mathbf{H}$ and predicted labels $\hat{\mathbf{Y}}$ with $f_\theta$ and $f_\phi$ by Eq.~\eqref{repre} and Eq.~\eqref{equ: pred_head} 
\IF{$num\_epoch$ \% $t$ = 0}
	\STATE Obtain two sets of counterfactuals $\left\{\mathcal{G}_i^{c_k} \mid k=1, \ldots, K\right\}$ and $\left\{\mathcal{G}_i^{e_k} \mid k=1, \ldots, K\right\}$ with Eq.~\eqref{equ: select cf1} and \eqref{equ: select cf2}.
\ENDIF
\STATE Compute the overall objective $\mathcal{L}$ with Eq.~(\ref{equ: all})
\STATE Update $\theta$ and $\phi$ according to the objective $\mathcal{L}$
\ENDFOR
\RETURN $f_\theta$ and $f_\phi$
\end{algorithmic}
\end{algorithm}

\section{Experiments}

In this section, we conduct experiments to evaluate the effectiveness of the proposed method and compare it with state-of-the-art fair GNNs. Specifically, we aim to answer the following questions: 
\begin{itemize}[leftmargin=*]
    \item \textbf{(RQ 1)} How effective is the proposed {\method} for fair node classification task on both synthetic datasets and real-world datasets?
    \item \textbf{(RQ 2)} Can the proposed {\method} find appropriate counterfactuals? 
    \item \textbf{(RQ 3)} How do the proposed modules work? How can each regularization term affect the model performance?
\end{itemize}

\begin{table*}[t]
\centering
\caption{Node classification performance and group fairness performance on real-world datasets.}
\label{tab: real-world}
\vskip -1em
\begin{tabular}{c|c|ccc|cc|cc|c}
\hline \textbf{Dataset} & \textbf{Metrics} & \textbf{GCN} & \textbf{GraphSAGE} & \textbf{GIN} & \textbf{FairGNN} & \textbf{EDITS}  & \textbf{NIFTY} &  \textbf{GEAR} & \textbf{{{\method}}} \\
\hline  & AUC ($\uparrow$) &  \underline{\facc{74.00}{1.51}}	 & \textbf{\facc{74.54}{0.86}} & \facc{72.69}{1.02} & \facc{65.85}{9.49}  & \facc{69.76}{5.46}  & \facc{72.05}{2.15} & \facc{65.80}{3.00} & \facc{71.87}{1.33} \\
& F1 ($\uparrow$) & \facc{80.05}{1.20} & \facc{81.15}{0.97} & \textbf{\facc{82.62}{1.55}} & \underline{\facc{82.29}{0.32}} & \facc{81.04}{1.09}  & \facc{79.20}{1.19} & \facc{78.04}{2.07} & \facc{82.16}{0.22} \\
\textbf{German} & $\Delta_{S P}$ ($\downarrow$) &  \facc{41.94}{5.52} & \facc{23.79}{6.70} & \facc{14.85}{4.64} & \underline{\facc{7.65}{8.07}} & \facc{8.42}{7.35}   & \facc{7.74}{7.80}    & \facc{8.60}{3.47}  & \textbf{\facc{6.60}{1.66}}  \\
& $\Delta_{E O}$ ($\downarrow$) &  \facc{31.11}{4.40} & \facc{15.13}{5.74} & \facc{8.26}{6.72}  & \underline{\facc{4.18}{4.86}} & \facc{5.69}{2.16}   & \facc{5.17}{2.38}    & \facc{6.34}{2.31}  & \textbf{\facc{1.58}{1.14}} \\
\hline \multirow{5}{*}{ \textbf{Bail} } & AUC ($\uparrow$) & \facc{88.50}{1.80} & \facc{90.50}{2.10} & \facc{77.30}{6.90} & \facc{88.20}{3.50} & \facc{89.07}{2.26}  & \textbf{\facc{92.04}{0.89}}  & \facc{89.60}{1.60} & \underline{\facc{91.39}{0.34}} \\
& F1 ($\uparrow$) &  \facc{78.20}{2.30} & \underline{\facc{80.40}{3.20}} & \facc{65.60}{8.40} & \facc{78.40}{2.10} & \facc{77.83}{3.79}  & \facc{77.81}{6.03}  & \facc{80.00}{3.10} & \textbf{\facc{83.09}{0.98}} \\
& $\Delta_{S P}$ ($\downarrow$) &  \facc{7.50}{1.40}  & \facc{8.60}{3.90}  & \facc{6.50}{3.40}  & \facc{7.40}{2.60}  & \underline{\facc{3.74}{3.54}}    & \facc{5.74}{0.38}   & \facc{5.80}{1.70}  & \textbf{\facc{2.29}{1.06}}  \\
& $\Delta_{E O}$ ($\downarrow$) &  \facc{2.30}{1.90}  & \facc{3.90}{2.20}  & \facc{4.10}{2.30}  & \facc{4.60}{1.30}  & \facc{4.46}{3.50}   & \facc{4.07}{1.28}   & \underline{\facc{1.90}{2.30}}  & \textbf{\facc{1.17}{0.52}} \\
\hline  \multirow{5}{*}{ \textbf{Credit} } & AUC ($\uparrow$) &  \facc{68.40}{1.90} & \textbf{\facc{75.60}{1.10}} & \facc{70.60}{1.00} & \facc{68.00}{2.10} & \underline{\facc{75.04}{0.12}} & \facc{72.89}{0.44} & \facc{74.00}{0.80} & \facc{73.42}{1.89} \\
& F1 ($\uparrow$) &  \facc{79.40}{2.70} & \facc{82.10}{0.80} & \facc{80.50}{1.60} & \facc{78.00}{4.20}  & \facc{82.41}{0.52}  & \facc{82.60}{1.25} & \underline{\facc{83.50}{0.80}} & \textbf{\facc{83.63}{0.89}} \\
& $\Delta_{S P}$ ($\downarrow$) &  \facc{10.80}{3.10} & \facc{10.90}{3.00} & \facc{13.00}{3.70} & \facc{18.70}{3.60}  & \facc{11.34}{6.36} & \facc{10.65}{1.65}  & \underline{\facc{10.40}{1.30}} & \textbf{\facc{8.63}{2.13}}  \\
& $\Delta_{E O}$ ($\downarrow$) &  \facc{8.70}{3.50}  & \facc{9.40}{3.30}  & \facc{12.10}{4.20} & \facc{17.50}{3.50}   & \facc{9.38}{5.39} & \underline{\facc{8.10}{1.91}}   & \facc{8.60}{1.80}  & \textbf{\facc{6.85}{1.55}}\\
\hline
\multicolumn{2}{c|}{\textbf{Avg. (Rank)}} & 5.58 & 4.42 & 5.75 & 5.92 & 4.67 & \underline{3.83} & 4.08 & \textbf{1.75} \\
\hline
\end{tabular}
\end{table*}

\begin{table}[t]
\centering
\caption{Real-world dataset statistics.}
\label{tab:dataset}
\vskip -1em
\small
\begin{tabular}{lccc}
\hline \text { Dataset } &\text { German Credit } & \text { Bail } & \text { Credit Defaulter } \\
\hline \text { \# Nodes }  & 1,000 & 18,876 & 30,000 \\
\text { \# Edges }  & 22,242 & 321,308 & 1,436,858 \\
\text { \# Attributes }  & 27 & 18 & 13 \\
\text { Sens. }  & \text { Gender } & \text { Race } & \text { Age } \\
\text { Label }  & \text { Credit status } & \text { Bail decision } & \text { Future default } \\
\hline
\end{tabular}
\vskip -1em
\end{table}

\subsection{Experiment Settings}

\subsubsection{Real-World Datasets} 
We conduct experiments on three widely used real-world datasets, namely German Credit~\citep{asuncion2007uci}, Credit Defaulter~\citep{yeh2009the}, Bail~\citep{jordan2015the}. The statistics of the datasets can be found in Table~\ref{tab:dataset}. %
The details of the datasets are as follows: 
\begin{itemize}[leftmargin=*]
    \item \textbf{German Credit}~\citep{asuncion2007uci}: the nodes in the dataset are clients and two nodes are connected if they have high similarity of the credit accounts. The task is to classify the credit risk level as high or low with the sensitive attribute ``gender''.
    \item \textbf{Credit Defaulter}~\citep{yeh2009the}: the nodes in the dataset are used to represent the credit card users and the edges are formed based on the similarity of the payments information. The task is to classify the default payment method with sensitive attribute ``age''.
    \item \textbf{Bail}~\citep{jordan2015the}: these datasets contain defendants released on bail during 1990-2009 as nodes. The edges between two nodes are connected based on the similarity of past criminal records and demographics. The task is to classify whether defendants are on bail or not with the sensitive attribute "race". 
\end{itemize}

\subsubsection{Synthetic Dataset} Real-world datasets do not offer ground-truth counterfactuals, prompting us to construct a synthetic dataset based on the Structural Causal Model (SCM) as depicted in Figure~\ref{fig:scm}.
The primary advantage of a synthetic dataset is that it provides us with ground-truth counterfactuals for each node, which enables us to assess the quality of the obtained counterfactuals.
In our approach, we consider settings with binary sensitive attributes and binary labels. A graph with 2000 nodes is sampled in our implementation. To generate the desired counterfactuals, we maintain the same sampled value of noise variables and use consistent causal relationships for each node. 
The sensitive attributes and labels are sampled from two different Bernoulli distributions, with $s_i \sim \mathcal{B}(p)$ and $y_i \sim \mathcal{B}(q)$, respectively. This results in generating vectors $\mathbf{s}_i = [(s_i)_{\times N}]$ and $\mathbf{y}_i = [(y_i)_{\times N}]$.
Next, environment and content features, $\mathbf{e}_i$ and $\mathbf{c}_i$, are sampled from normal distributions $\mathbf{e}_i \sim \mathcal{N}(\mathbf{s}_i, \mathbf{I})$ and $\mathbf{c}_i \sim \mathcal{N}(\mathbf{y}_i, \mathbf{I})$, respectively. These features are combined to form the overall latent feature $\mathbf{z}_i = [\mathbf{c}_i , \mathbf{e}_i]$.
The observed feature for each node $v_i$, denoted as $\mathbf{x}_i$, is computed as $\mathbf{x}_i = \mathbf{W}\mathbf{z}_i + \mathbf{b}_i$, where $\mathbf{W}_{ij} \sim \mathcal{N}(1, 1)$, and $\mathbf{W} \in \mathbb{R}^{d_2 \times 2d_1}$, with $\mathbf{b}_{i} \sim \mathcal{N}(0,\mathbf{I}) \in \mathbb{R}^{d_2}$.
The adjacency matrix $\mathbf{A}$ is defined such that $\mathbf{A}_{ij} = 1$ if $\sigma(\operatorname{cos}(\mathbf{z}_i, \mathbf{z}_j) + \epsilon_{ij}) \geqslant \alpha$ and $i \neq j$, with $\epsilon_{ij} \sim \mathcal{N}(0,1)$, and $\mathbf{A}_{ij} = 0$ otherwise. Here, $\sigma(\cdot)$ denotes the Sigmoid function, and the threshold $\alpha$ controls the edge number.
We have the freedom to set sensitive attribute probability $p$, label probability $q$, latent feature dimension $2d_1$, observed feature dimension $d_2$, node number $N$, and threshold $\alpha$ to control the biased graph generation process.
Note that in the SCM we have $C \rightarrow Y$ instead of $Y \rightarrow C$, thus a better way is to first generate content features and then assign labels to the features. Intuitively, we argue that when using an optimal classifier to deal with content features with different means will assign the same label in our generation process. Therefore, to simplify the generation process, we use $C \rightarrow Y$ in our dataset design.

The synthetic dataset comes with notable advantages. Firstly, it gives us access to exact counterfactuals. After generating the initial graph, we keep all noise variables and unrelated variables unchanged, then adjust the sensitive attribute $s_i$ or label $y_i$ to calculate the precise counterfactual through the same graph generation procedure. Secondly, the synthetic dataset enables adjustable bias levels, providing us control over the extent of bias in our models. 
As a result, we can undertake a comprehensive and detailed evaluation of our model's fairness and prediction quality.

\subsubsection{Baselines}
To evaluate the effectiveness of {\method}, we include representative and state-of-the-art methods, which can be categorized into three categories: (1) \textit{plain node classification methods}: GCN~\citep{kipf2017semisupervised}, GraphSAGE~\citep{hamilton2017inductive} and GIN~\citep{xu2019how}; (2) \textit{fair node classification methods}: FairGNN~\citep{dai2021say}, EDITS~\citep{dong2022edits}; (3) \textit{graph counterfactual fairness methods}: NIFTY~\citep{agarwal2021towards} and GEAR~\citep{ma2022learning}.  Unless otherwise specified, we use GraphSAGE as the model backbone except for baseline GCN and GIN. We use SAGE to denote GraphSAGE. The detailed descriptions about the datasets are as follows:

\begin{itemize}[leftmargin=*]
    \item GCN~\cite{kipf2017semisupervised}: GCN is a popular spectral GNN, which adopts a localized first-order approximation of spectral graph convolutions.
    \item GraphSAGE~\cite{hamilton2017inductive}: GraphSAGE is a method for inductive learning that leverages node feature information to generate unsupervised embeddings for nodes in large graphs, even if they were not included in the initial training. 
    \item GIN~\cite{xu2019how}: Graph Isomorphism Network (GIN) is a graph-based neural network that can capture different topological structures by injecting the node's identity into its aggregation function.
    \item FairGNN~\cite{dai2021say}: FairGNN uses adversarial training to achieve fairness on graphs. It trains the learned representation via an adversary which is optimized to predict the sensitive attribute.
    \item EDITS~\cite{dong2022edits}: EDITS is a pre-processing method for fair graph learning. It aims to debias the input network to remove the sensitive information in the graph data.
    \item NIFTY~\cite{agarwal2021towards}: It simply performs a flipping on the sensitive attributes to get counterfactual data. It regularizes the model to be invariant to both factual and counterfactual data samples.
    \item GEAR~\cite{ma2022learning}: GEAR is a method for counterfactual fairness on graphs. It utilizes a variational auto-encoder to synthesize counterfactual samples to achieve counterfactual fairness for graphs.
\end{itemize}

\subsubsection{Evaluation Metrics}
We evaluate the model performance from three perspectives: classification performance, group fairness and counterfactual fairness. (\textbf{i}) For \textit{classification performance}, we use AUC and the F1 score to measure node classification performance. %
(\textbf{ii}) For \textit{fairness}, following~\cite{dai2021say}, we adopt two commonly used group fairness metrics, i.e., statistical parity (SP) $\Delta_{S P}$ and equal opportunity (EO) $\Delta_{E O}$, which are computed as $\Delta_{S P}=|P(\hat{y}_u=1 \mid s=0)-P(\hat{y}_u=1 \mid s=1)|$ and $\Delta_{E O}=|P(\hat{y}_u=1 \mid y_u=1, s=0)-P(\hat{y}_u=1 \mid y_u=1, s=1)|$. The smaller $\Delta_{SP}$ and $\Delta_{E O}$ are, the fairer the model is. (\textbf{iii}) For \textit{counterfactual fairness}, as we have the ground-truth counterfactuals on the synthetic dataset, Following~\citep{ma2022learning}, we use the counterfactual fairness metric $\delta_{C F}$, i.e., $\delta_{C F}=|P\left(\left(\hat{y}_i\right)_{S \leftarrow s} \mid \mathbf{X}, \mathbf{A}\right)-P\left(\left(\hat{y}_i\right)_{S \leftarrow s^{\prime}} \mid \mathbf{X}, \mathbf{A}\right)|$, 
where $s, s^\prime \in \{0,1\}^N$ are the sensitive attributes and $s^\prime = 1 - s$. $(\hat{y}_i)_{S \leftarrow s^{\prime}} $ is the computed ground-truth counterfactual label with the same data generation process as shown in Figure~\ref{fig:scm}. We use subscript $S \leftarrow s^\prime$ to denote counterfactual computation~\citep{shanmugam2018elements}, i.e., keeping the same data generation process and values of random noise variable. Counterfactual fairness of the graph is only measured on synthetic dataset.

\subsubsection{Setup}
For German Credit, Credit Defaulter and Bail
, we follow train/valid/test split in~\citep{agarwal2021towards}. 
For the constructed synthetic dataset, we use a 50/25/25 split for training/validation/testing data. We randomly initialize the parameters. 
For each combination of the hyper-parameters configuration, we run the experiments with 10 random seeds and grid search for the best configuration.

\subsection{Performance Comparison}

To answer RQ1, we conduct experiments on real-world datasets and synthetic dataset with comparison to baselines. 

\subsubsection{Performance on Real-World Datasets}
Table~\ref{tab: real-world} shows the average performance with standard deviation of ten runs on real-world datasets. The best results are highlighted in \textbf{bold} and the runner-up results are \underline{underlined}. From Table~\ref{tab: real-world}, we observe: 
\begin{itemize}[leftmargin=*]
    \item {\method} can improve the group fairness performance. Across three datasets, Table~\ref{tab: real-world} shows {\method} can make fairer predictions than other baseline methods. {\method} beats all the baselines with respect to the group fairness metrics.
    \item There exists a trade-off between group fairness and prediction performance. Plain node classification methods, such as GCN, GraphSAGE and GIN, tend to have better prediction performance and worse group fairness performance. Fair node classification methods, including FairGNN, EDITS, NIFTY, GEAR and {\method}, tend to suffer from a prediction performance drop and the group fairness performance is better.
    \item {\method} achieves best performance on the prediction-fairness trade-off. We use the average rank of two prediction metrics and two group fairness metrics to know the performance of the trade-off. Our model ranks 1.75 and the runner-up model ranks 3.83. 
    Our model outperforms the state-of-the-art node representation learning methods, which shows the effectiveness of our model.
    \item Graph counterfactual fairness methods, such as NIFTY, GEAR and {\method}, achieved better performance than other baselines. Correlation-based counterfactual notions can capture the causal relationships and help to boost the group fairness performance. 
\end{itemize}

\begin{figure}[t]
\centering
  \begin{subfigure}[b]{0.23\textwidth}
        \centering
        \includegraphics[height=1in]{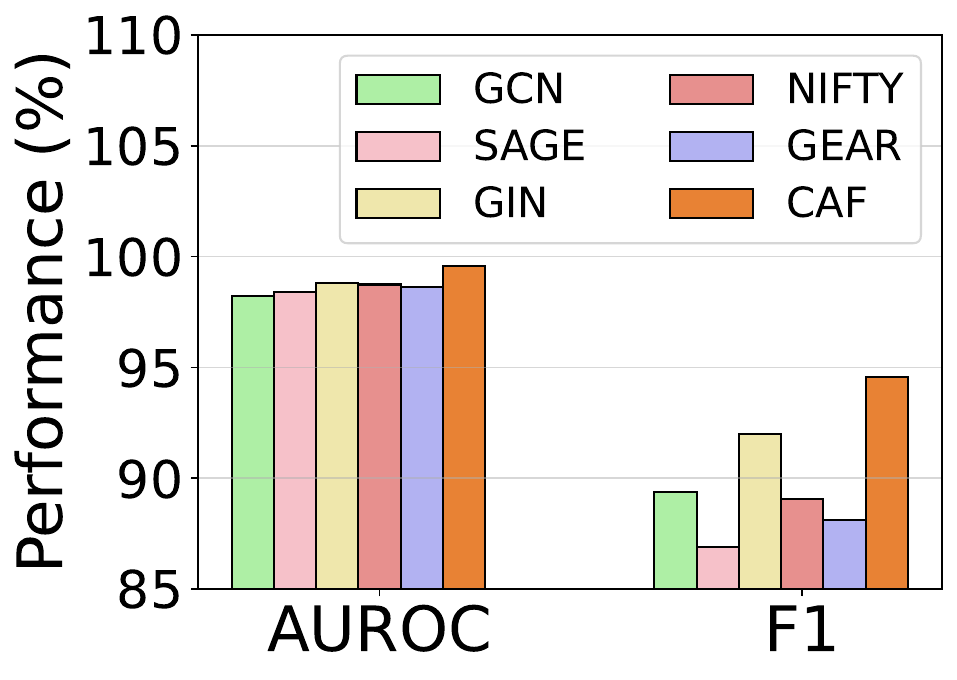}
        \caption{Prediction}
    \end{subfigure}
  \begin{subfigure}[b]{0.23\textwidth}
        \centering
        \includegraphics[height=1in]{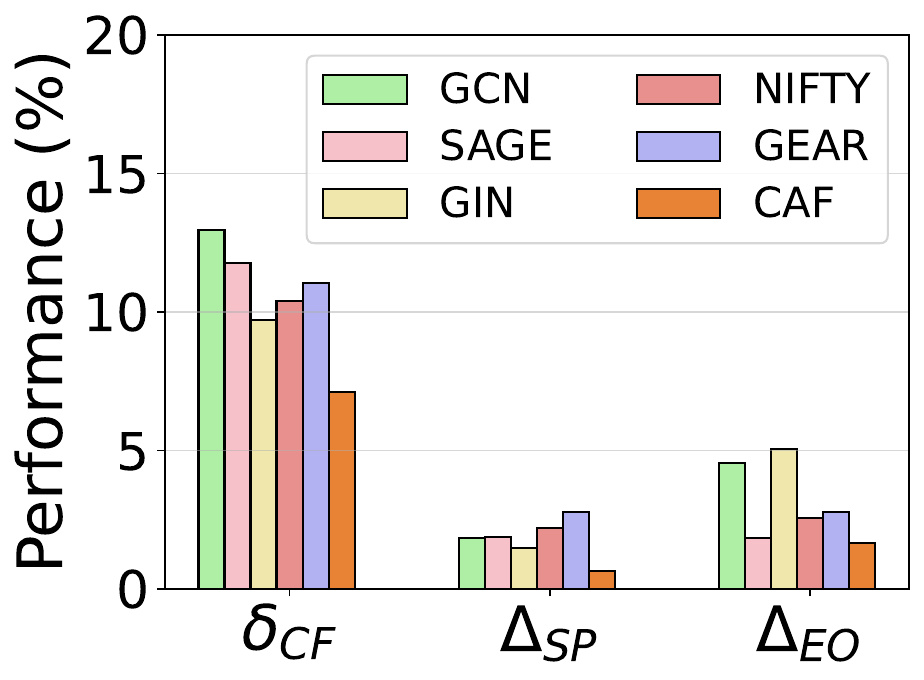}
        \caption{Fairness}
    \end{subfigure}
    \vspace{-1em}
  \caption{Node classification performance and group fairness performance on synthetic datasets.
  }
  \label{fig: synthetic}
\end{figure}

\subsubsection{Performance on Synthetic Dataset}
Figure~\ref{fig: synthetic} reports the performance on the synthetic dataset. %
On the synthetic dataset, we have the desired ground-truth counterfactuals, which can be used to measure the performance of graph counterfactual fairness. We compare our model with plain node classification models and counterfactual fairness models. The observations are as follows:
\begin{itemize}[leftmargin=*]
    \item  {\method} beats all the models with respect to the prediction, group fairness and counterfactual fairness metrics. We argue that in our assumed biased generation process, our model can effectively find invariant, sufficient and informative representations to make accurate and fair predictions.
    \item Other graph counterfactual fairness-based methods, including NIFTY and GEAR, cannot consistently outperform other methods. These methods design their model without considering meaningful causal relationships. NIFTY simply perturbs the sensitive attribute and omits the further influence on features and graph structure. GEAR adopts an GraphVAE to model the causal relationships, which may fail to generate meaningful counterfactuals.
\end{itemize}

\begin{figure}[t]
\centering
  \begin{subfigure}[b]{0.23\textwidth}
        \centering
        \includegraphics[height=1.3in]{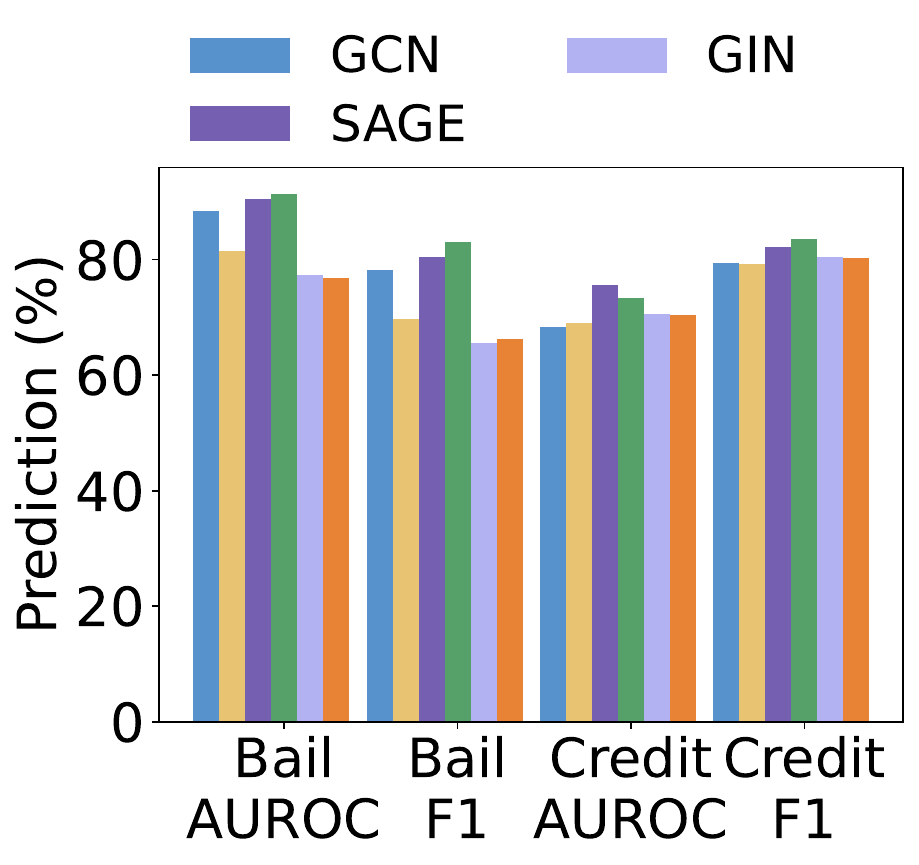}
        \caption{Prediction}
    \end{subfigure}
  \begin{subfigure}[b]{0.23\textwidth}
        \centering
        \includegraphics[height=1.3in]{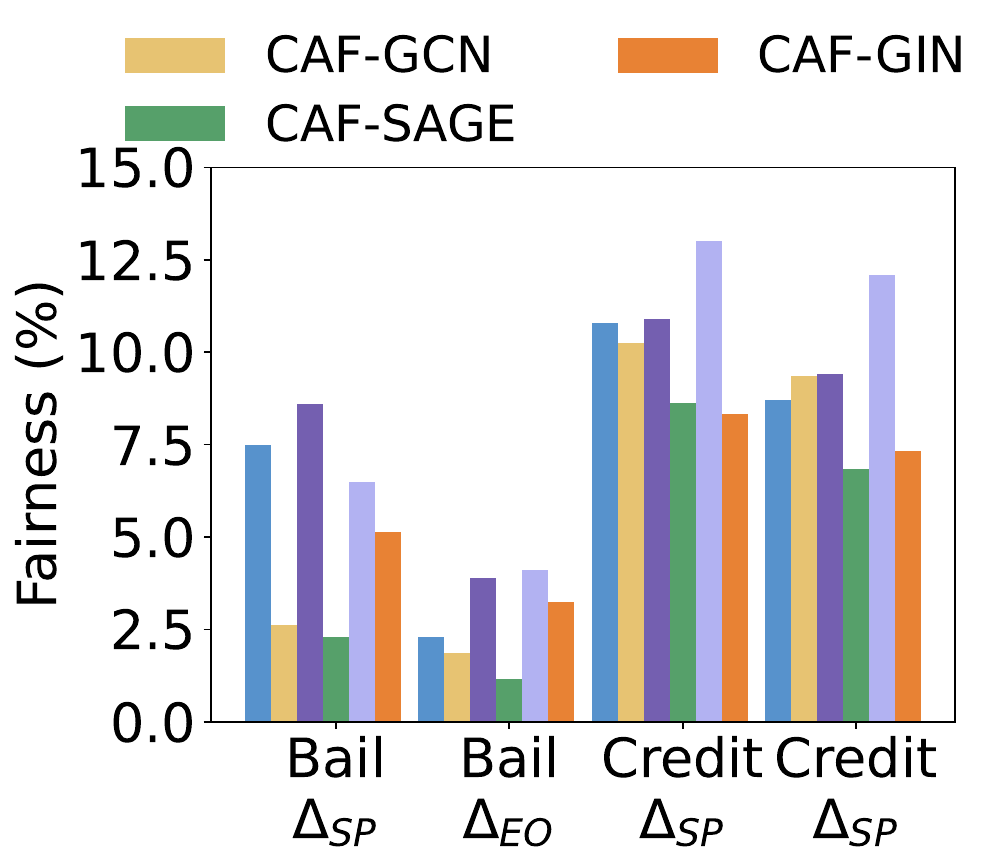}
        \caption{Fairness}
    \end{subfigure}
    \vskip -1em
  \caption{Comparison of the prediction performance and fairness performance of different backbones.}
  \label{fig: backbone}
  \vspace{-1em}
\end{figure}

\subsection{Flexibility of {\method} for Various Backbones}
To show the flexibility of {\method} in improving the fairness of various backbones while maintaining high classification accuracy, other than GraphSAGE, we also plug our model in GCN and GIN. Figure~\ref{fig: backbone} shows the classification performance and fairness performance on Bail and Credit. From Figure~\ref{fig: backbone}, we observe that compared with the backbones, {\method} can significantly improve the fairness with no or marginal decrease in classification performance. For example, 
on Bail dataset, the prediction performance with GIN backbone drops by 0.54\% on AUROC but the $\Delta_{SP}$ drops by 1.37\% and the $\Delta_{EO}$ drops by 0.86\%, which is an improvement on fairness performance.
This demonstrates the flexibility of {\method} in benefiting various backbones.  %

\subsection{Quality of Counterfactuals}

\begin{figure}[t]
    \centering
    \includegraphics[width=0.26\textwidth]{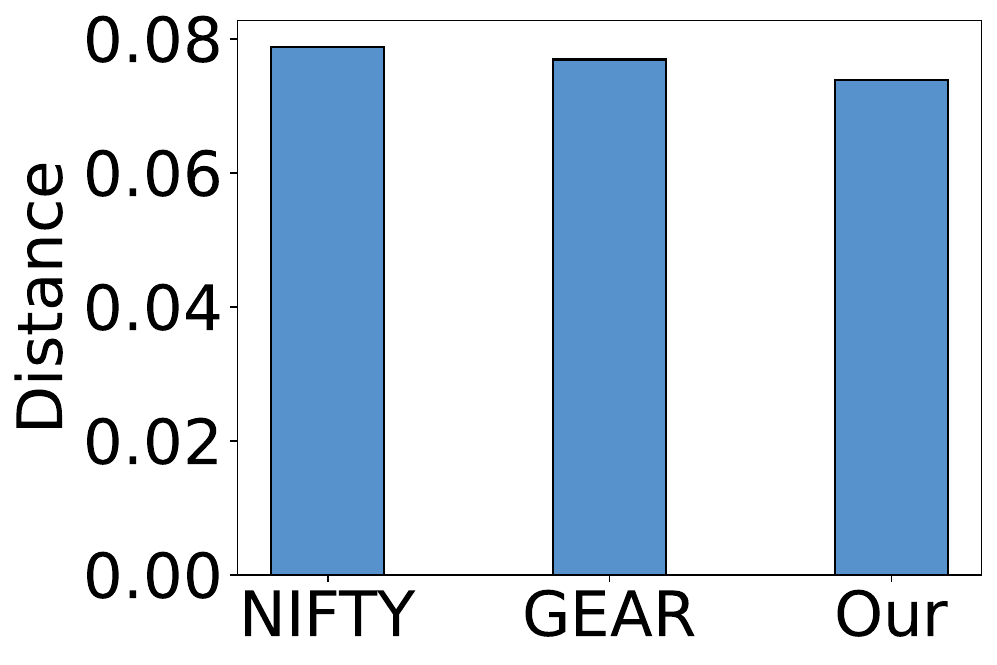}
    \vskip -1em
    \caption{ Discrepancy between learned counterfactual representation and ground-truth counterfactual representation.
    }
    \label{fig:case}
    \vskip -1em
\end{figure}

To answer RQ2, we compare the counterfactuals obtained by {\method} with ground-truth counterfactuals to investigate whether we can obtain the desired counterfactuals. We conduct experiments on the synthetic dataset which has ground-truth counterfactuals. 
We first use {\method} to obtain counterfactuals. To measure the discrepancy of the obtained counterfactuals with respect to the feature and structure in the ego graph, we compare the learned counterfactual representations and the ground-truth counterfactual representations.
We compare our model with two graph counterfactual fairness baselines, i.e., NIFTY~\citep{agarwal2021towards} and GEAR~\citep{ma2022learning}. NIFTY simply flips the sensitive attributes to get their counterfactuals. GEAR uses a GraphVAE to generate the counterfactuals based on self-perturbation and neighbor-perturbation. Figure~\ref{fig:case} shows the average result for all the nodes on the synthetic dataset. We show that {\method} can find better counterfactuals than other graph counterfactual fairness models, i.e., smaller discrepancy to ground-truth counterfactuals. The result also shows there is still space for existing methods to improve the performance of getting appropriate counterfactuals. 

\subsection{Ablation Study}
\label{ablation}

In our model, the pre-trained model can provide pseudo-labels for the nodes in the unlabeled set. 
Thus, we can select counterfactuals from the entire dataset. 
The model trained from scratch, without any pre-training, is denoted as {\method-NP}. Without pseudo-labels, we can only select counterfactuals from the training set, which is denoted as the variant {\method-NS}. We evaluate the performance on  synthetic dataset. The results are reported in Table~\ref{tab: ablation}. We find the model {\method-NS} performs worse than the {\method} but better than {\method-NP}. The result shows the pseudo-labels can also boost the performance of our model. Usually, the training set is small and the model may not obtain desired counterfactuals from the limited data points. Although pseudo-labels may contain some noisy information, they can also help improve the our model performance. 

We further delve into how the constraints impact performance. When merely setting $\alpha=0$ or $\beta=0$, we denote the model as {\method-NA} and {\method-NB}, respectively. The models {\method-NA} and {\method-NB} outperform {SAGE}, yet fall short when compared to {\method}. This indicates that both the sufficiency and invariance constraints collectively contribute to the superior performance of our model.

\begin{table}[t]
\centering
\caption{Ablation study.}
\label{tab: ablation}
\vskip -1em
\resizebox{.45\textwidth}{!}{%
\begin{tabular}{cccccc}
\toprule
\multirow{2}*{\textbf{Models}} & \multicolumn{4}{c}{\textbf{Synthetic Dataset}}  \\
\cline{2-6}
 & \textbf{AUROC}  $\uparrow$ & \textbf{F1}  $\uparrow$ & $\mathbf{\delta_{CF}}$  $\downarrow$ &$\mathbf{\Delta_{DP}}$  $\downarrow$ & $\mathbf{\Delta_{EO}}$  $\downarrow$  \\
\toprule 
\text{SAGE} & \facc{98.40}{0.31} & \facc{86.89}{3.59} & \facc{11.76}{2.92} & \facc{1.87}{0.82} & \facc{1.85}{1.41} \\
\text{\method} & \facc{99.57}{0.06} & \facc{94.58}{0.58} & \facc{7.12}{0.41}  & \facc{0.65}{0.41} & \facc{1.66}{0.57} \\
\text{\method}-NA & \facc{99.33}{0.27} & \facc{94.72}{2.15} & \facc{10.42}{1.98} & \facc{1.63}{0.41} & \facc{1.83}{0.74}  \\
\text{\method}-NB & \facc{98.47}{0.28} & \facc{88.64}{4.31} & \facc{8.31}{3.22}  & \facc{0.91}{0.28} & \facc{1.63}{0.51} \\
\text{\method}-NP & \facc{98.36}{0.39} & \facc{84.72}{3.07} & \facc{13.47}{2.62} & \facc{1.37}{0.55} & \facc{1.96}{1.78}  \\
\text{\method}-NS & \facc{98.81}{0.73} & \facc{91.98}{3.25} & \facc{9.72}{2.05}  & \facc{1.35}{0.18} & \facc{1.73}{0.55} \\
\bottomrule
\end{tabular}%
}
\vskip -2em
\end{table}

\begin{figure}[t]
\centering
  \begin{subfigure}[b]{0.22\textwidth}
        \centering
        \includegraphics[height=1.2in]{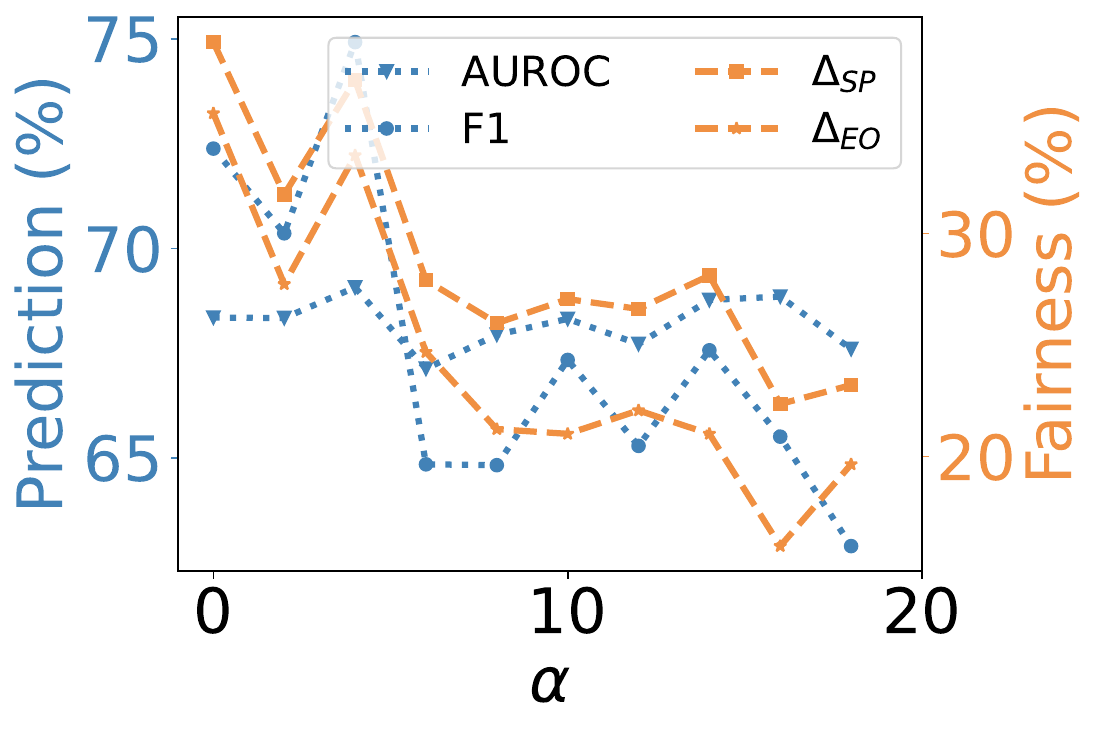}
    \end{subfigure}
  \begin{subfigure}[b]{0.22\textwidth}
        \centering
        \includegraphics[height=1.2in]{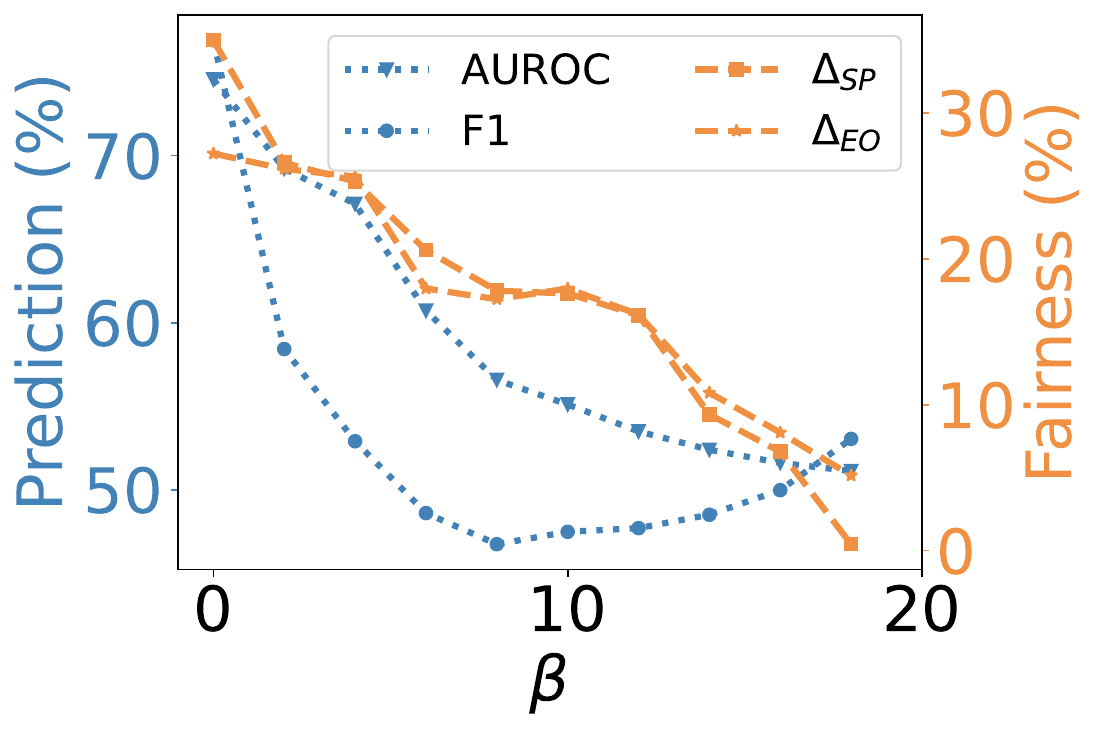}
    \end{subfigure}
    \vspace{-1.2em}
  \caption{Hyper-parameter study on German dataset.}
  \label{fig: hyper}
  \vspace{-2em}
\end{figure}

\subsection{Hyper-Parameter Sensitivity Analysis}
There are two important hyperparameters in {\method}, i.e., $\alpha$ and $\beta$. $\alpha$ controls the contribution of the invariance regularization $\mathcal{L}_{\text{inv}}$ and $\beta$ controls the contribution of the sufficiency regularization. To understand the impact of $\alpha$ and $\beta$ on {\method}, we fix $\beta$ as 5 and vary $\alpha$ as $\{0, 1,\dots, 18\}$. Similarly, we fix $\alpha$ as 1 and vary $\beta$ as $\{0, 1,\dots, 18\}$. 
We report the result on German dataset in Figure~\ref{fig: hyper}. From Figure~\ref{fig: hyper}, we have the following observations: there exists a trade-off between prediction performance and fairness performance. The trend is that when we increase the $\alpha$ and $\beta$, we will get worse prediction performance and better fairness performance. We argue that without these regularizations, the model may rely on sensitive information. When we decrease the regularization budget, we can disentangle content representations. Thus, the prediction performance will get worse and the fairness performance will be better.

\section{Conclusion and Future Work}
In this paper, we study the problem of learning fair node representations with GNNs. We first formalize the biased 
graph generation process with an SCM. Motivated by causal theory, we propose a novel framework {\method} to learn fair node presentations which meet graph counterfactual fairness criteria and can achieve good prediction-fairness performance. Specifically, we align the model design with the data generation process and convert the problem to learn content representations. 
We derive several properties of the optimal content representation from the causal graph, i.e., invariance, sufficiency and informativeness. To get appropriate supervision for the invariance regularization, we design a counterfactual selection module. 
Extensive experiments demonstrate that {\method} can achieve state-of-the-art performance on synthetic dataset and real-world datasets with respect to the prediction-fairness trade-off. 

There are several interesting directions worth exploring. First, in this paper, we mainly focus on binary classification and binary sensitive attribute. We will extend the work to multi-class classification and multi-category sensitive attributes. Second, in this paper, we focus on static graphs while there are many different kinds of graphs in real-world. Thus, we aim to extend our model to more complex graph learning settings, such as dynamic graphs, multi-value sensitive attributes and labels.

\begin{acks}
This material is based upon work supported by, or in part by the National Science Foundation (NSF) under grants number IIS-1909702, IIS-2153326, and IIS-2212145, Army Research Office (ARO) under grant number W911NF-21-1-0198, Department of Homeland Security (DHS) CINA under grant number E205949D, and Cisco Faculty Research Award.  
\end{acks}

\bibliographystyle{ACM-Reference-Format}
\balance
\bibliography{a}

\end{document}